\documentclass{article}

\usepackage[nonatbib,final]{neurips_2023}

\usepackage[utf8]{inputenc} 
\usepackage[T1]{fontenc}    
\usepackage{hyperref}       
\usepackage{url}            
\usepackage{booktabs}       
\usepackage{amsfonts}       
\usepackage{nicefrac}       
\usepackage{microtype}      
\usepackage{xcolor}         

\usepackage{graphicx}
\usepackage{comment}

\title{MHG-GNN: Combination of Molecular Hypergraph Grammar with Graph Neural Network}

\author{%
  Akihiro Kishimoto \\
  IBM Research - Tokyo\\
  \And
  Hiroshi Kajino \\
  IBM Research - Tokyo\\
  \And
  Masataka Hirose \\
  JSR Corporation \\
  \And
  Junta Fuchiwaki \\
  JSR Corporation \\
  \And  
  Indra Priyadarsini \\
  IBM Research - Tokyo\\
  \And    
  Lisa Hamada \\
  IBM Research - Tokyo\\
  \And
  Hajime Shinohara \\
  IBM Research - Tokyo\\
  \And
  Daiju Nakano \\
  IBM Research - Tokyo\\
  \And
  Seiji Takeda \\
  IBM Research - Tokyo\\
}
\begin{document}

\maketitle

\begin{abstract}
  Property prediction plays an important role in material discovery. 
  As an initial step to eventually develop a foundation model for material science, 
  we introduce a new autoencoder called the MHG-GNN, which combines graph neural network (GNN) with Molecular
  Hypergraph Grammar (MHG).
 Results on a variety of property prediction tasks with diverse materials show that MHG-GNN is promising. 
\end{abstract}

\section{Introduction} \label{sec:intro}

Machine learning for effective material design has been a topic of interest.
Machine learning models for accurately predicting chemical or physical properties help material scientists have better understandings to the materials of interest more quickly than simulations and trial-and-error experiments. 

A major obstacle is the limited availability of training data in material science. 
A prediction task for a specific target property often has only tens of pairs of molecular structures and their target values \cite{Takeda:KDD2020}.
Data-hungry, modern machine learning algorithms cannot benefit from only such a small dataset. 

A \emph{foundation model}~(FM)~\cite{Bommasani:arxiv2021} trained with a huge unlabeled dataset aims to address the data shortage. With a small labeled dataset and the FM, each prediction task is modeled as a \emph{downstream} task. Because of public, large-scale unlabeled \emph{base} datasets on molecular structures~\cite{Irwin:JCIM2020,Kim:NAR2021}, building FMs for materials science~\cite{Ahmad:Arxiv2022, Takeda:AAAI2023} has attracted much attention.

One approach discussed in~\cite{Takeda:AAAI2023} requires the FMs for materials science to meet at least the following three requirements.
First, \emph{any} molecule should be embedded into a latent vector representation.
Second, the latent representation should effectively represent features of molecules to be able to address a wide variety of downstream tasks including property prediction.
Third, a decoder included for molecular optimization should map a latent vector into a \emph{structurally valid} molecule \emph{without any failure}. 
Satisfying these requirements allows the FM to solve many property prediction tasks with little effort as well as to directly achieve an ultimate goal of optimizing molecules. 

As a first step toward this end, we present MHG-GNN, an autoencoder architecture
that has an encoder based on GNN and a decoder based on a sequential model with MHG~\cite{Kajino:ICML2019}.
Since the encoder is a GNN variant, MHG-GNN can accept any molecule as input, and  
demonstrate high predictive performance on molecular graph data.
In addition, the decoder inherits the theoretical guarantee of MHG on always generating a structurally valid molecule as output.

As initial experiments, we evaluate MHG-GNN with a total of six downstream tasks on three different kinds of materials,
and show that MHG-GNN performs better than two other well-known approaches, while opening up further opportunities as a next step
to develop the FM for material science.

\section{Background} \label{sec:background}

We review the work related to our attempt to
develop a unimodal model of MHG-GNN supporting three requirements in Section~\ref{sec:intro}, 
aiming to be extended to an approach discussed in~\cite{Takeda:AAAI2023} in the future.

Since SMILES is a text format representing a two-dimensional molecular graph, molecules are regarded either as sequences of characters or as graphs (see \cite{Faez:Access2021} for a survey).
Irrespective of whether to pretrain models or not, existing approaches perform learning without any labels. 

Although the requirements we have discussed include a valid decoder, decoders are not often supported in both representations \cite{Ahmad:Arxiv2022, Anstine:JACM2023, Rong:NeurIPS2020, Ross:NMI2022, Ying:NeurIPS2021, Yueksel:MLSC2023, Yun:NeurIPS2019}. 
Even if the decoders are available, many of them do not handle structural constraints such as the total number of bonds for a carbon, occasionally resulting in invalid molecules violating these constraints \cite{Shepherd:Arxiv2020, Gomez:ACS2018,Honda:Arxiv2019,Kipf:BDL2016,Ma:NeurIPS2018,Simonovsky:ICANN2018}.

Large language models\cite{Anil:Arxiv2023, OpenAI:Arxiv2023,Touvron:Arxiv2023} based on Transformer \cite{Vaswani:NeurIPS2017} and its variants \cite{Devlin:ACL2019, Radford:2018}
attempt to learn efficient latent representations on natural language.  
There are attempts to deal with SMILES as language e.g., \cite{Chithrananda:Arxiv2020, Honda:Arxiv2019}. 
However, in addition to the decoder issues, misclassification in the latent space may occur
if two similar molecules have very different SMILES strings \cite{Jin:ICML2018}. 

There are approaches to support valid decoders such as 
JT-VAE \cite{Jin:ICML2018}, MHG-VAE \cite{Kajino:ICML2019} and SELFIES \cite{Krenn:MLSC2020}.  
However, for a new molecule, JT-VAE can utilize only a part of the latent vector due to its junction tree representation. 
MHG-VAE may not be able to encode it to a latent vector.
From a viewpoint of downstream tasks related to one of the requirements,
the latent vector of any molecule must be computed to be able to assess the molecules of interest. 

A SELFIES-based decoder can be combined with our approach. However, as a first step, we focus on an MHG-based decoder.  
A comparison with the SELFIES-based decoder remains future work.

\section{MHG-GNN}

To the best of our knowledge, in designing autoencoder \cite{Kingma:ICLR2014}, all previous approaches represent their input and output in the \emph{identical} format such as from SMILES string to SMILES string \cite{Gomez:ACS2018, Honda:Arxiv2019}. 
Using the identical format ends up with only partially resolving the issues discussed so far. 

MHG-GNN is an autoencoder with \emph{different} representations as its input and output. This leads
to meeting all the requirements to be a part of the FM.  
MHG-GNN receives a two-dimensional graph as input to its GNN-based encoder, and the output of its RNN-based decoder is a production rule sequence of MHG.
Unlike MHG-VAE, MHG-GNN can encode any molecular graph without accounting for the existence of the production rules in the base dataset as well as to directly embed graph structures into their latent space. 
Additionally, like MHG-VAE, the decoder of MHG-GNN can always return structurally valid molecules stemming from MHG.

\subsection{Encoder}

Our encoder employs a Graph Isomorphism Network (GIN) \cite{Xu:ICLR2019} variant that additionally accounts for embedding edges \cite{Hu:ICLR2020, Liu:ICLR2022}.

Each atom of a molecule is represented as chemical features, e.g., atomic number, formal charge, and aromaticity. 
Each atom feature is embedded to a vector with the same dimension size as its corresponding node in GIN.
All embedded atom features are summed to represent an initial vector $h_i^{0}$ for GIN node $i$. 
Each edge of a molecular structure such as bond type is embedded analogously to an embedding vector $e_{j,i}^{0}$
on its GIN undirected edge between nodes $j$ and $i$.

In iteration $k$, the encoder performs so-called message passing for each node $i$, defined as follows: 
$$
h_i^{k+1} = \mbox{MLP}((1 + \epsilon) h_i^k + \sum_{j \in N(i)}\mbox{ReLU}(h_j^{k} + e_{j,i}))
$$

\noindent where $N(i)$ is a set of direct neighbors of $i$, $\epsilon$ is a trainable parameter, $\mbox{MLP}$ is a neural network module, and $\mbox{ReLU}$ is a Rectified Linear Unit. 
In our encoder, $\mbox{MLP}(x):=\mbox{Lin}(\mbox{ReLU}(\mbox{BN}(\mbox{Lin}(x))))$, where
$\mbox{Lin}$ is a linear layer and \mbox{BN} is a one-dimensional Batch Normalization layer \cite{Ioffe:ICML2015}. 

Finally, the entire representation $h_G$ of graph $G$ is used as a latent vector to address downstream tasks.
As suggested in \cite{Xu:ICLR2019} from a theoretical viewpoint on model expressiveness, 
$h_G$ is defined as: 
$$
h_G= \mbox{CONCAT}(\sum_{i \in V_G} h_i^{k} | k = {0, 1, \cdots,r})
$$
\noindent where \mbox{CONCAT} concatenates vectors, $V_G$ is a set of nodes in $G$, and $r$ is the maximum iteration size. 

\subsection{MHG}

\begin{figure}[t]
  \begin{center}
    \includegraphics[scale=0.5]{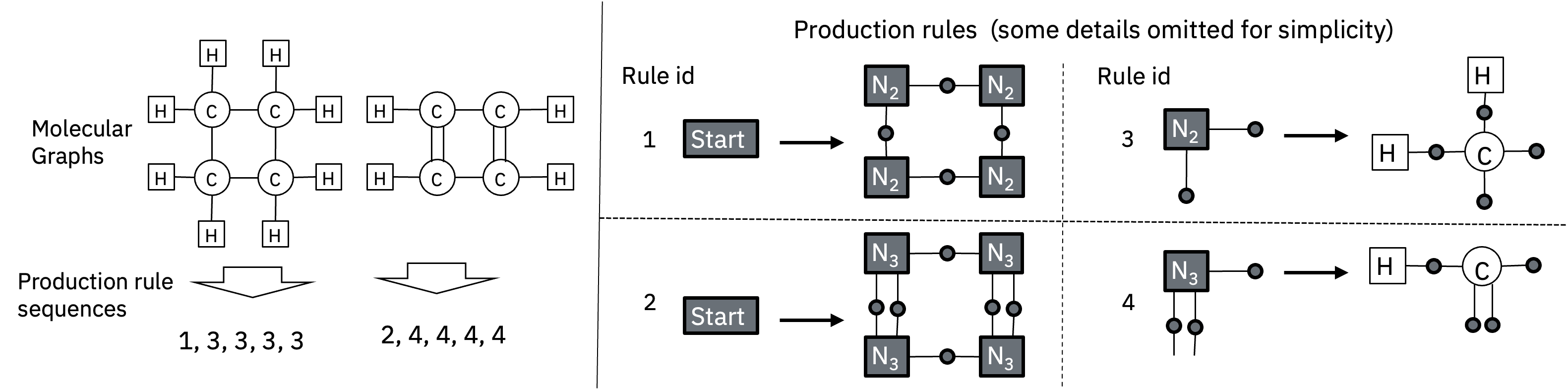}
  \end{center}
  \caption{MHG and production rule sequence. Note implicit hydrogen is used in practice. } \label{fig:mhg}
\end{figure}

Approaches available to always generate valid molecular structures include MHG \cite{Kajino:ICML2019}, Junction Tree (JT) \cite{Jin:ICML2018} and Reversible JT (RJT) \cite{Ishitani:JCIM2022}. 
We choose MHG for MHG-GNN, because MHG enables MHG-GNN to build a simpler architecture as well as to leverage the entire embedding.

Given a dataset on molecular structures, MHG generates production rules on molecular hypergraphs.
Figure \ref{fig:mhg} shows an example of four rules extracted from two molecules.  
Each rule has a precondition and an effect defined at the left and right of its arrow, respectively.  
An edge of a normal molecular graph is split into two parts connected with one small, gray circle called a \emph{hypernode} in its corresponding hypergraph.
A \emph{hyperedge} denoted as a square is connected to any number of hypernodes.
A square with an atom is called a terminal symbol. A non-terminal symbol includes a non-atom symbol. 

For a molecule $M$, MHG attempts to find a sequence of production rules to generate $M$,
starting with the \emph{start} symbol. 
In applying rule $r$ to hypergraph $G$, MHG checks whether $G$ has a subgraph that matches the precondition of $r$ or not.
If $G$ has such a subgraph, that subgraph is replaced with the effect of $r$.
In this rule application, any hypernode in $G$ continues preserving exactly two hyperedges, which is essential to ensure the structural validity. 
If $G$ does not have any subgraph to match, $r$ is safely ignored. This leads to significantly
reducing the number of production rules to consider.

If $G$ has only terminal symbols, $G$ is easily converted to a structurally valid molecule by just removing all hypernodes from $G$. Otherwise, MHG needs to find an applicable rule that can further reduce $G$.  

\subsection{Decoder}

While $h_G$ is used as a latent vector, additional layers are prepared before it is processed by the decoder. 
$h_{G}$ is passed to a mean module and a log variance module on multi-dimensional Gaussian distributions to perform
the reparameterization trick \cite{Higgins:ICLR2017}. 
Each of these modules performs $\tanh(\eta\mbox{Lin}(h_G))$, where $\eta$ is a trainable scalar. 
After its output is adapted with a linear layer, it is passed to the initial hidden unit of the decoder. 

The decoder employs an architecture similar to that of MHG-VAE \cite{Kajino:ICML2019}. 
It consists of layers of Gated Recurrent Unit (GRU) \cite{Cho:EMNLP2014}
and handles as output a sequence of production rule embeddings.
Each rule embedding in the sequence is converted back to a probability constrained by MHG as necessary.

\section{Experimental Results}

Our base dataset consists of 1,381,747 molecules extracted from the PubChem database \cite{Kim:NAR2021}.
16,362 rules are generated to represent these molecules as production rule sequences.

\subsection{Setup for Downstream Tasks}

\begin{table}[!t]
  \caption{Summary on datasets}\label{tab:dataset}
  \begin{center}
    {\small 
\begin{tabular}{|c||c|c|c|c|}
  \hline
  Dataset name & Size & Property value creation & Structural variety & Miscellaneous \\
  \hline
  \hline
  Polymer & 5000 & Simulations & High & Monomer weight $\leq$ 300 \\
  \hline
  Photoresist & 7884 & Simulations & High & \\
  \hline
  Chromophore & 346 & Experiments & Low & \\
  \hline  
\end{tabular}
}
\end{center}
\end{table}

Table \ref{tab:dataset} shows our three proprietary datasets containing materials of diverse characteristics.

We evaluate the following methods that perform six downstream tasks on various property predictions:  
(1) {\bf $\mbox{MHG-GNN}_r$} with a radius of $r=3, 5, 6, 7, 8$ (i.e., the number of iterations for neighborhood aggregations) and $256(r+1)$ dimensions, (2) {\bf ECFP6}, a well-known fingerprint descriptor on two-dimensional molecular substructures with 1024 bit \cite{Rogers:JCIM2010}, and (3) {\bf Mordred},  a freely-available, high-performing software calculating 1825 two- and three- dimensional descriptors \cite{Moriwaki:JC2018}. We exclude three-dimensional descriptors of Mordred for a fairer comparison to the others that consider only two-dimensional structures. This results in Mordred using 1613 dimensions.

Since $r$ is regarded as a hyperparameter, which value should be used needs to be decided before the test dataset is evaluated.  We select $r$ based on the best $R^2$ score on the validation dataset. 

In our preliminary experiments with a different dataset, we find that MHG-VAE cannot encode roughly $50\%$ of the molecules due to a failure of generating appropriate production rule sequences. Our evaluation, therefore, excludes MHG-VAE here. 

We split these downstream datasets into training, validation and test datasets with the ratio of 0.6, 0.2 and 0.2, respectively. We create downstream models by the LightGBM algorithm \cite{Ke:NeurIPS2017} with the RMSE loss function and optimize their hyperparameters by Optuna \cite{Akiba:KDD2019} using the validation datasets.

\subsection{Results} \label{sec:results}

\begin{table}[!t]
  \caption{Performance on various downstream tasks (test datasets, $R^2$ score)}\label{tab:results}
  \begin{center}
    {\small 
\begin{tabular}{|cc||c|c|c|c|c|c|}
  \hline
  & & \multicolumn{3}{c|}{Polymer} & \multicolumn{2}{c|}{Photoresist} & Chromophore \\
  \hline 
         &                      &                   &  Bulk             & Refractive        &                         &                   & $\lambda_{max}$ \\
  \multicolumn{2}{|c||}{Method} & Density           & modulus           & index             & HOMO                    & LUMO              & on NIR \\
  \hline
  \hline
  $\mbox{MHG-GNN}$ &            & {\bf 0.578}       & {\bf 0.516}       & {\bf 0.865}       & {\bf 0.896}             & {\bf 0.845}       &  {\bf 0.845}\\    
  \hline  
  ECFP6 &                       & 0.523             & 0.482             & 0.823             & 0.791                   & 0.782             &  0.708 \\
  \hline    
  Mordred &                     & 0.567             & 0.505             & 0.859             & 0.894                   & 0.830             &  0.842\\
  \hline  
\end{tabular}
}
\end{center}
\end{table}

Table \ref{tab:results} shows $R^2$ scores calculated with the test datasets.
The best numbers are marked in bold. 
Our results clearly demonstrate that MHG-GNN outperforms ECFP6 and Mordred in all downstream tasks, and 
that MHG-GNN is an important candidate for property prediction tasks on material discovery.

The performance of MHG-GNN tends to be improved when its radius is increased
from 3 to 6 or 7 except bulk modulus. 
Its performance then tends to be worsened when the radius is further increased to 8.
One hypothesis to explain this behavior is that MHG-GNN might suffer from an essential issue that still remains open. It is notoriously known that diminishing returns is clearly observed when GNN is constructed with deeper layers \cite{Liu:2018}. 
The other hypothesis is that the radius of 6-7 (i.e., the diameter of 12-14) might be sufficient to capture most of the important structural information in our case.
Many of the molecules in the Polymer and Photoresist datasets contain two ring structures, while the Chromophore dataset has molecules with two or three rings. 

\section{Conclusions}

As a first step to build an FM, we have introduced MHG-GNN and evaluated its performance on various prediction tasks on diverse materials,  
showing that MHG-GNN outperforms Mordred and ECFP6, two of well-known approaches applicable to property prediction. 

There are numerous research directions as future work. For example, more comprehensive understanding to MHG-GNN is necessary, including comparisons with other pretrained models such as \cite{Honda:Arxiv2019, Jin:ICML2018, Ross:NMI2022}, integrations with other decoders including SELFIES, and evaluations with a larger base dataset as well as other downstream tasks
including molecular optimization.  
Improving the performance of MHG-GNN on downstream tasks is also of interest, such as development of new model architectures that overcome the performance degradation with many GNN layers.
Finally, toward a second step to build the FM, extending MHG-GNN to support multimodalities (e.g., three-dimensional coordinates and properties easily calculated) is an important topic to investigate.

\bibliography{paper}
\bibliographystyle{abbrv}

\newpage

\appendix

\section{Supplementary Material}

\subsection{Configurations of MHG-GNN}

We implement MHG-GNN in Python using PyTorch \cite{Pytorch}, PyTorch Geometric (PyG) \cite{PyG} and RDKit \cite{RDKit},  
and train its models with the $\beta$-VAE loss function \cite{Higgins:ICLR2017},
the Adam optimizer \cite{Kingma:ICLR2015} and the ReduceLROnPlateau scheduler on a cluster of machines consisting of Intel E5-2667 CPUs at 3.30GHz and NVIDIA A100 Tensor Core GPUs. 

As defined in PyG, each atom and bond consist of 9 and 3 features, respectively, and are embedded to 256 dimensions for a node in GIN. 
The dimension size of Gaussian distributions is also set to 256. 
The values of $\epsilon$ and $\eta$ in MHG-VAE are initialized to 1 and 0.
Each GRU layer consists of 384 hidden units and its layer size is set to 3. 
The production rule embedding size is set to 128.

To train MHG-GNN, we set the batch size, the dropout rate, the frequency of scheduler invocations, $\lambda$ for Adam, and $\beta$ for $\beta$-VAE to $512$, $0.1$, $1000$, $5.0 \times 10^{-4}$,
and $0.01$, respectively.

\subsection{Detailed Performance}

\begin{table}[!t]
  \caption{Performance on various downstream tasks (test datasets, $R^2$ score)}\label{tab:details}
  \begin{center}
    {\small 
\begin{tabular}{|cc||c|c|c|c|c|c|}
  \hline
  & & \multicolumn{3}{c|}{Polymer} & \multicolumn{2}{c|}{Photoresist} & Chromophore \\
  \hline 
         &                      &                   &  Bulk             & Refractive        &                         &                   & $\lambda_{max}$ \\
  \multicolumn{2}{|c||}{Method} & Density           & modulus           & index             & HOMO                    & LUMO              & on NIR \\
  \hline
  \hline
  $\mbox{MHG-GNN}_3$ & (1024)   & 0.578 & 0.523             & 0.863             & 0.884                   & 0.835             &  0.793\\
  \hline
  $\mbox{MHG-GNN}_5$ & (1536)   & 0.503             & {\bf 0.528}       & 0.863             & 0.890                   & 0.838             &  0.828 \\  
  \hline
  $\mbox{MHG-GNN}_6$ & (1792)   & {\bf 0.588}       & 0.516 & 0.863             & 0.885                   & 0.845 &  0.832 \\
  \hline
  $\mbox{MHG-GNN}_7$ & (2048)   & 0.577             & 0.522             & {\bf 0.866}       & {\bf 0.896} & {\bf 0.848}       &  {\bf 0.845}\\
  \hline
  $\mbox{MHG-GNN}_8$ & (2304)   & 0.566             & 0.520             & 0.865 & 0.885                   & 0.841             &  0.813\\
  \hline
  \hline
  $\mbox{MHG-GNN}$ & (selected)   & 0.578             & 0.516             & 0.865 & 0.896                   & 0.845             &  0.845\\  
  \hline
  \hline  
  ECFP6 & (1024)                & 0.523             & 0.482             & 0.823             & 0.791                   & 0.782             &  0.708 \\
  \hline    
  Mordred & (1613)              & 0.567             & 0.505             & 0.859             & 0.894                   & 0.830             &  0.842\\
  \hline  
\end{tabular}
}
\end{center}
\end{table}

Table \ref{tab:details} comprehensively shows $R^2$ scores calculated with the test datasets. 
The size of the latent space is denoted next to each model name. 
The numbers showing the best performance are marked in bold.
The performance of  MHG-GNN selected is also shown (see Subsection \ref{sec:results}). 

\end{document}